\documentclass[]{spie}  

\usepackage{amsfonts}
\usepackage{amsmath}

\usepackage[utf8]{inputenc} 
\usepackage[T1]{fontenc}    
\usepackage{hyperref}       
\usepackage{url}            
\usepackage{booktabs}       
\usepackage{amsfonts}       
\usepackage{nicefrac}       
\usepackage{microtype}      
\usepackage{xcolor}         
\usepackage{bbm}
 
\usepackage{amsfonts}
\usepackage{graphicx}
\usepackage{amsfonts}
\usepackage{graphicx}
\usepackage{epsfig}
\usepackage{graphicx}
\usepackage{amsmath}
\usepackage{amssymb}
\usepackage{mathtools}
\usepackage{resizegather}
\usepackage{marvosym}
\usepackage{dsfont}
\usepackage{enumitem}
\usepackage{caption}
\usepackage{multirow}  
\usepackage{amsmath,amsfonts,amssymb}
\usepackage{graphicx}

\title{Fair Text to Medical Image Diffusion Model with Subgroup Distribution Aligned Tuning}

\author[a]{Xu Han}
\author[b]{Fangfang Fan}
\author[c]{Jingzhao Rong}
\author[a]{Zhen Li}
\author[a,c]{Georges El Fakhri}
\author[c]{Qingyu Chen}
\author[a,c]{Xiaofeng Liu}

\affil[a]{Department of Radiology and Biomedical Imaging, Yale University, New Haven, CT 06519, USA}
\affil[b]{Department of Neurology, Beth Israel Deaconess Medical Center, Harvard Medical School, Boston, MA 02140, USA}
\affil[c]{Department of Biomedical Informatics and Data Science, Yale University, New Haven, CT 06519, USA}

\authorinfo{Xu Han and Fangfang Fan contributed equally to this work\\Corresponding to: xiaofeng.liu@yale.edu}

\begin{document} 
\maketitle

\begin{abstract}
The Text to Medical Image (T2MedI) approach using latent diffusion models holds significant promise for addressing the scarcity of medical imaging data and elucidating the appearance distribution of lesions corresponding to specific patient status descriptions. Like natural image synthesis models, our investigations reveal that the T2MedI model may exhibit biases towards certain subgroups, potentially neglecting minority groups present in the training dataset. In this study, we initially developed a T2MedI model adapted from the pre-trained Imagen framework. This model employs a fixed Contrastive Language–Image Pre-training (CLIP) text encoder, with its decoder fine-tuned using medical images from the Radiology Objects in Context (ROCO) dataset. We conduct both qualitative and quantitative analyses to examine its gender bias. To address this issue, we propose a subgroup distribution alignment method during fine-tuning on a target application dataset. Specifically, this process involves an alignment loss, guided by an off-the-shelf sensitivity-subgroup classifier, which aims to synchronize the classification probabilities between the generated images and those expected in the target dataset. Additionally, we preserve image quality through a CLIP-consistency regularization term, based on a knowledge distillation framework. For evaluation purposes, we designated the BraTS18 dataset as the target, and developed a gender classifier based on brain magnetic resonance (MR) imaging slices derived from it. Our methodology significantly mitigates gender representation inconsistencies in the generated MR images, aligning them more closely with the gender distribution in the BraTS18 dataset.

  
\end{abstract}


\section{Introduction}

Latent diffusion models are at the forefront of text-to-image generation, having achieved remarkable success in creating high-quality images from textual descriptions \cite{saharia2022photorealistic,zhu2023conditional,li2024snapfusion}. Recent advancements have extended these capabilities to the domain of synthetic text-to-medical image generation (T2MedI) \cite{kim2024controllable,huang2024chest}. This innovation holds significant potential for augmenting medical image datasets with images that depict specific clinical scenarios as described by text prompts. Furthermore, integrating these synthetic images with clinical diagnostic reports can significantly enhance the utility of both resources, providing insights into the potential appearance distributions of lesions corresponding to detailed patient status descriptions \cite{kim2024controllable,huang2024chest}.

Despite these advances, conventional text-to-image models such as Imagen \cite{saharia2022photorealistic} often exhibit biases related to sensitive subgroups, including gender, race, and age \cite{shen2023finetuning}. While several diffusion debiasing techniques have been developed to address these biases, achieving absolute demographic parity (e.g., 50\% White and 50\% Asian representation) may not always align with the nuanced needs of specific contexts. Instead, a more contextually appropriate approach might aim for subgroup proportions that reflect those of the target dataset, recognizing that definitions of fairness can vary across different applications.

To our knowledge, this study is the first to address bias and fairness within text-to-medical image generation using cutting-edge diffusion models. We focus particularly on the sensitive factor of gender, which can be accurately classified from MR scans \cite{ebel2023classifying}. This paper provides both qualitative and quantitative analyses of gender bias in our T2MedI model. We also introduce an innovative Subgroup Distribution Aligned Tuning (SDAT) scheme, designed to effectively adjust the generated subgroup distributions to more closely match those of the target dataset, thus mitigating biases.

\begin{figure}[!t]
\begin{center}
\includegraphics[width=0.9\linewidth]{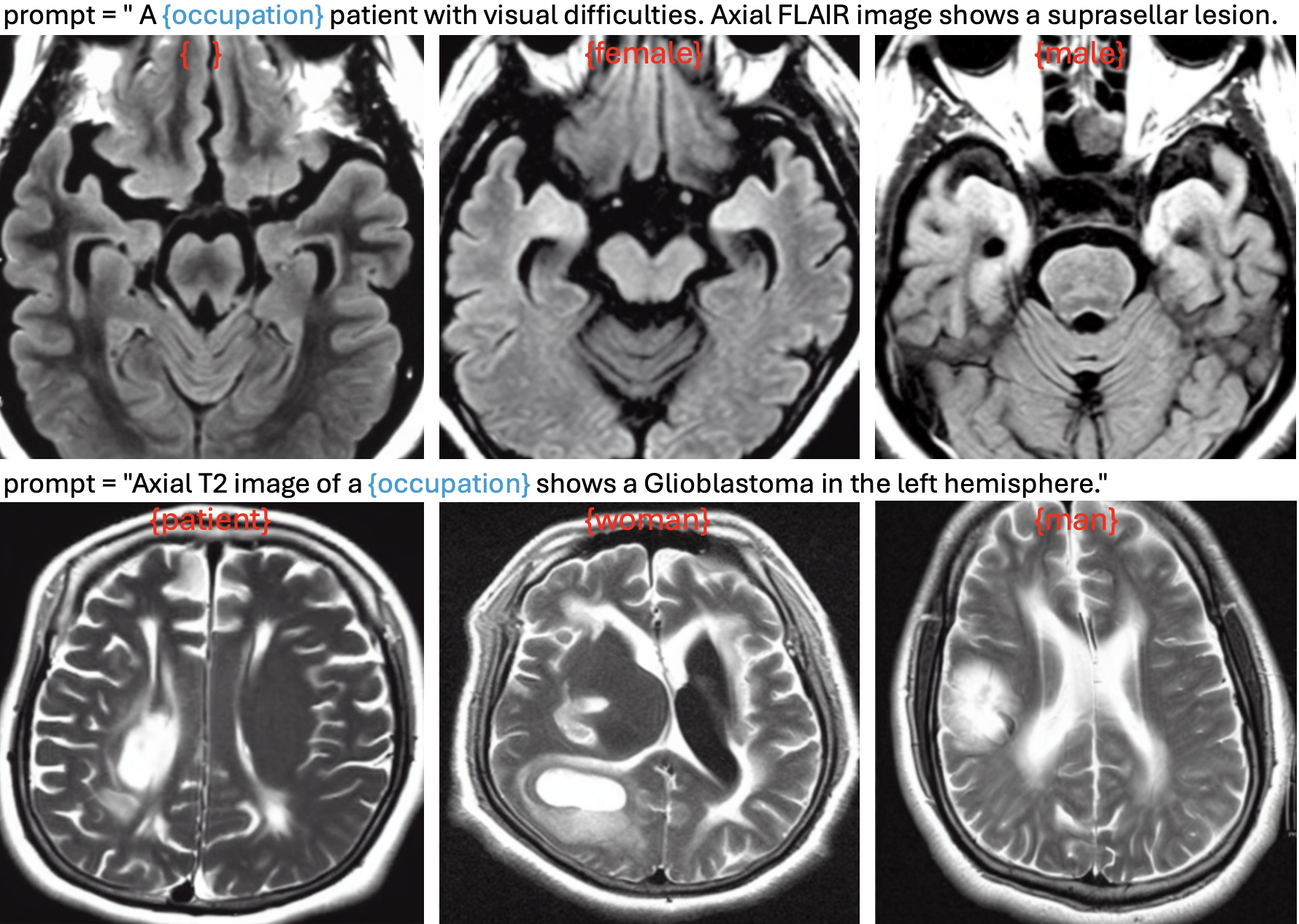}    
\end{center} 
\caption{Examples of T2MedI generated images without gender-related prompts, with female-related prompts, and with male-related prompts, respectively.} 
\label{fig:illus} 
\end{figure}

\section{METHODS}\vspace{+5pt}

\subsection{Text to Medical Image (T2MedI) Generation Model}

Our T2MedI model is initialized using the successful latent diffusion foundation of Imagen \cite{saharia2022photorealistic}, which is trained on a large dataset of natural images to generate photo-realistic outputs at a resolution of 256$\times$256. The core of its architecture, the noise prediction network $\epsilon_\theta$, receives inputs from the encoded text prompt $\phi(T_i)$, the encoded image representation $f(x_i)$, and the time step $t$. In our framework, the text encoder $\phi(\cdot)$ from CLIP ViT-L/14, as utilized in Imagen, remains fixed \cite{saharia2022photorealistic}. We have fine-tuned the $f(\cdot)$ function and diffusion modules using paired text prompts and images from the Radiology Objects in Context (ROCO) dataset, comprising 81,000 images \cite{pelka2018radiology}, to specialize the model in radiological imaging.

\subsection{Subgroup Distribution Aligned Tuning (SDAT)}

Addressing subgroup consistency in generated images poses a distribution alignment challenge \cite{shen2023finetuning}. We focus on managing sensitive factors among $K$ subgroups in T2MedI-generated images to align with a desired target dataset's subgroup distribution. In each fine-tuning iteration, we generate a batch of MR slices $\{x_i\}_{i=1}^N$ using the current diffusion model with varied text prompts $\{T_i\}_{i=1}^N$. Each generated MR slice $x_i$ is then assessed by a high-performance subgroup classifier that produces a softmax output to predict the probability distribution of each subgroup $p_i=\{p_i^k\}_{k=1}^K$.

Concurrently, we sample a batch of independent and identically distributed (i.i.d.) samples $\{x^t_i\}_{i=1}^N$ from the target dataset and analyze another set of softmax histograms $u_i\in\mathbb{R}^K$ using the same subgroup classifier. 

When the target dataset's subgroup labels are known, we can construct a fixed sample set reflecting a specific subgroup proportion (e.g., a balanced gender distribution with half men and half women). The optimal transport between the generated $\{p_i\}_{i\in[N]}$ and target distributions $\{u_{\sigma_i}\}_{i\in[N]}$ is determined by solving for $\sigma^*$ in the following equation:
{\begin{equation}
\begin{aligned}
\sigma^* =\underset{\sigma\in P_N}{\operatorname{argmin}} \sum_{i=1}^N |p_i-u_{\sigma_i}|,\label{eq:0}
\end{aligned}
\end{equation}}where $P_N$ represents all permutations of $N$ samples, $\sigma=\{\sigma_i\}_{i=1}^N$ and $\sigma_i\in[N]$. This solution identifies the most efficient modification needed for the current images to match the target distribution \cite{shen2023finetuning}. The expectation $q_i=\mathbb{E} \{u_{\sigma^*}\}, \forall i\in [N]$ forms a probability vector where each element $k$ represents the likelihood of image $x_i$ belonging to target subgroup $k$. The alignment is enforced through cross-entropy loss $\mathcal{L}{align}$ applied only to predictions exceeding a confidence threshold $\tau$:
\begin{align}
\mathcal{L}{align}=\frac{1}{B}\sum_{i=1}^B \mathbbm{1} [c_i \geq \tau] \mathcal{L}_{CE}(p_i,y_i)
\end{align}Therefore, only the relatively confident predictions are used to calculate $\mathcal{L}_{align}$ with corresponding $\mathcal{L}_{CE}(p_i,y_i)$.

To mitigate catastrophic forgetting of the radiological image generation capabilities during fine-tuning, we employ a knowledge distillation strategy. We use a copy of the frozen, non-fine-tuned T2MedI model to guide image generation. This approach includes an image semantics preserving regularization term $\mathcal{L}{reg}$:
{\begin{equation}
\begin{aligned}
\mathcal{L}{reg}=\frac{1}{N}\sum_{i=1}^N[1-\cos(\text{CLIP}(x_i), \text{CLIP}(o_i))]
\end{aligned}
\end{equation}}where $\{o_i\}_{i=1}^N$ is a batch of generated images with the original T2MedI diffusion model. This term penalizes dissimilarity between each pair of fine-tuned and original generated images $x_i$ and $o_i$, which are conditioned on the same initial noise in the diffusion model. The fine-tuning process incorporates both $\mathcal{L}{align}$ and $\mathcal{L}{reg}$ to ensure faithful image generation.

 \section{Experiments and Results}

To evaluate our model, we focused on gender representation in the generated T2-weighted MR images, distinguishing between two subgroups: male and female ($K = 2$). We utilized a prompt template "Axial T2 image of a glioma patient showing ${\text{medical condition}}$," with 100 medical conditions used for training and 20 for testing, such as "moderate mass effect," "a supra-sellar lesion," and "hyperintensity in the left mesencephalon." Each prompt was used to generate 100 images with varying noise initializations through our T2MedI diffusion model.

\subsection{Quantitative Evaluation:}

For quantitative analysis, we adopted two different target gender distributions: an even split of 50\% male and 50\% female, and a skewed distribution of 70\% male to 30\% female. These distributions were selected to mirror potential real-world demographics in clinical settings. The gender classifier, built on an 18-layer ResNet architecture, was trained on the BraTS18 dataset, which includes MR scans from 158 female and 211 male subjects. For a balanced gender distribution (50\% female), we selected 158 subjects from each gender. For the skewed distribution (30\% female), we chose 90 female and 210 male subjects.

We quantified bias using the following metric:
\begin{align}
    bias(T)=\frac{1}{(k-1)K/2}\sum_{i,j\in[K]:i<j}|freq(i)-freq(j)|,
\end{align}
where $freq(i)$ and $freq(j)$ represent the frequency of group $i$ or $j$ in the generated or target images, respectively. This metric evaluates the deviation from a balanced representation within the generated images.

From Table 1, it is evident that our Subgroup Distribution Aligned Tuning (SDAT) approach significantly reduces bias in both cases of target gender proportions. This demonstrates the effectiveness of SDAT in producing more equitable gender representation in the synthesized MR images, aligning closely with the predefined target distributions.

\subsection{Qualitative Evaluation:}
Figure 1 illustrates that prompts devoid of specific gender indicators predominantly generate MR images with characteristics similar to those typically observed in females.

\begin{table}[t!]
\centering  
\caption{Numerical comparisons of gender bias without or with our proposed SDAT.}\vspace{+5pt}
\resizebox{0.7\linewidth}{!}{
\begin{tabular}{l|c|c}
\hline
Method & Bias (target 50\% female) $\downarrow$ & Bias (target 30\% female) $\downarrow$\\\hline
w/o SDAT & 0.513  &  0.748  \\ 
\hline
with SDAT & 0.214  &  0.353  \\
\hline
\end{tabular}\label{tab1}}  
\end{table}




\section{Discussion and CONCLUSION}

In this study, we addressed gender bias in the T2MedI model, a diffusion model-based text-to-medical image generation system. We introduced an efficient Subgroup Distribution Aligned Tuning (SDAT) scheme that utilizes a pre-trained gender classifier and an optimal transport-based alignment loss to adjust the gender proportion in the generated images. Additionally, we implemented a regularization term to mitigate catastrophic forgetting, thus preserving the quality of generated images. Our approach demonstrated substantial effectiveness in aligning generated images with actual gender distributions in the BraTS18 dataset.

Despite these positive results, the extent of bias mitigation can vary significantly across different applications, highlighting the context-dependent nature of our model's application. Furthermore, the biases inherent in the underlying language model, although not the primary focus of this study, represent a critical area for future research. Such biases need to be addressed to prevent skewed data representation before practical implementation.

It is also noteworthy that the generated medical images might include artifacts or ``hallucinations" that could potentially lead to misinterpretations of patient status. Consequently, we recommend that all generated images be subject to thorough review by experienced clinicians prior to clinical use. This practice will ensure the reliability and safety of using AI-enhanced diagnostic tools in real-world medical settings.

Our results, validated on the BraTS18 dataset \cite{bakas2018identifying}, demonstrate a significant reduction in bias, showcasing the potential of our method to be adopted in diverse medical imaging applications. The presentation will also delve into the inherent challenges of language model biases and their impact on medical image generation, suggesting avenues for further research. We would also highlight the necessity of integrating clinical oversight in the deployment of AI-driven tools in healthcare to safeguard against misinterpretations that could affect patient care.


\vspace{+5pt}
\acknowledgments 

This work is supported by NSF NAIRR240016, NIH R21EB034911, and Google Cloud research credits.


\bibliography{main} 
\bibliographystyle{spiebib} 

\end{document}